# A Preliminary Exploration of the Differences and Conjunction of Traditional PNT and Brain-inspired PNT

Xu He[1], Xiaolin Meng[1]*, Wenxuan Yin[1], Youdong Zhang[1], Lingfei Mo[1]*, Xiangdong An[1], Fangwen Yu[2], Shuguo Pan[1], Yufeng Liu[1], Jingnan Liu[3], Yujia Zhang[1], Wang Gao[1]

[1] School of Instrument Science and Engineering, State Key Laboratory of Comprehensive PNT Network and Equipment Technology, Southeast University, Nanjing 210096, China

[2] Center for Brain Inspired Computing Research, Department of Precision Instrument, Tsinghua University, Beijing 100084, China

[3] Research Center of Satellite Navigation and Positioning Technology, Wuhan University, Wuhan 430079, China

*Corresponding author(s). E-mail(s): xiaolin_meng@seu.edu.cn; lfmo@seu.edu.cn
Contributing authors: hexu@seu.edu.cn; wenxuan_yin@seu.edu.cn; ydzhang@seu.edu.cn; xiangdong.an@seu.edu.cn; yufangwen@tsinghua.edu.cn; psg@seu.edu.cn; yufeng_liu@seu.edu.cn; jnliu@whu.edu.cn; yujia_zhang@seu.edu.cn; gaowang1990@seu.edu.cn

**Abstract**

Developing universal Positioning, Navigation, and Timing (PNT) is our enduring goal. Today's complex environments demand PNT that is more resilient, energy-efficient and cognitively capable. This paper asks how we can endow unmanned systems with brain-inspired spatial cognition navigation while exploiting the high precision of machine PNT to advance universal PNT. We provide a new perspective and roadmap for shifting PNT from "tool-oriented" to "cognition-driven". Contributions: (1) multi-level dissection of differences among traditional PNT, biological brain PNT and brain-inspired PNT; (2) a four-layer (observation-capability-decision-hardware) fusion framework that unites numerical precision and brain-inspired intelligence; (3) forward-looking recommendations for future development of brain-inspired PNT.

# 1. Introduction

Unmanned system Positioning, Navigation, and Timing (PNT) technologies have achieved numerous practical advances. Particularly noteworthy is the rapid maturation of Global Navigation Satellite System (GNSS)-based PNT, which has not only expanded its application domains but also driven down operational costs. However, these technologies still face formidable challenges in highly uncertain and complex scenarios, such as deep space, the deep ocean, polar regions, and dense urban environments. These scenarios demand far more stringent requirements on PNT resilience. In 2018, Yang et al. [1] identified ten pivotal scientific challenges that humanity must tackle within the next decade. Among these, navigation and exploration in complex, unknown environments were explicitly highlighted, underscoring the issue as a pressing frontier problem for the global scientific community.

For many years, we have endeavored to develop a universal PNT system that can function effectively in complex environments, one that is ubiquitously adaptable, robust, and offers controllable precision with high reliability. Yet the road to such a universal PNT is long and arduous, marked by formidable obstacles. Specifically, when deployed in challenging settings, machine PNT systems often confront GNSS-denied conditions and a lack of dependable prior knowledge or cooperative aids. In dynamic scenes, these systems are susceptible to motion blur. Under conditions of weak illumination or low visibility, they struggle to extract environmental features. Moreover, in confined spaces, they cannot effectively utilize re-localization to mitigate accumulated errors. Moreover, current machine PNT systems are inherently limited by insufficient robustness, weak adaptability, excessive power consumption, and a deficit in spatial cognition. Notably, Professor Fei-Fei Li, often referred to as the "godmother of Artificial Intelligence (AI)," has initiated a venture focused on spatial intelligence, aiming to endow large language models with spatial awareness [2]. This capability is precisely the cornerstone of brain-inspired PNT research and is indispensable to the advancement of universal PNT.

It is well known that animals in nature possess extraordinary navigation intelligence. Over billions of years of evolution and natural selection, their brains have developed spatial cognition and navigation capabilities that are fundamental to their survival. For instance, biologists who have tracked humpback whales during their migrations, which span thousands of kilometers, have documented the whales' routes and determined that their directional error across hundreds of kilometers is only about $1°$ [3]. Similarly, monarch butterflies, which weigh barely a hundred grams, undertake multi-generational migrations that span hundreds of kilometers [4]. Neuroscientists have discovered numerous active social place cells in bat brains that facilitate cooperative behavior and spatial navigation [5]. Eagles exhibit a highly developed 3D spatial cognition that enables them to navigate seamlessly across aerial, terrestrial and aquatic domains.

The innate spatial cognition intelligence has long been a focal point in neuroscience. In 2016, Finkelstein et al. [6] proposed a perspective for the neural circuitry underlying navigation, suggesting that the cooperative interactions among various navigation-related cell types across different brain regions can be likened to a Kalman filter. Specifically, place cells in the hippocampus function as the brain's Global Positioning System (GPS), relaying positional information to the entorhinal cortex. Head-direction cells in the parahippocampal region serve as a compass, providing directional cues. Speed cells in the entorhinal cortex act as an odometer, integrating motion signals, while grid cells fuse path integration cues to estimate the brain's pose and predict its trajectory. By integrating these internal and external spatial cues, the brain can anticipate current directional and positional errors and uses them to refine the brain's perception of "where I am" and "where I am going." These neurobiological insights have catalyzed the emergence of brain-inspired navigation, an interdisciplinary frontier that seeks to develop universal PNT systems by emulating the brain's spatial cognition prowess.

It is worth noting that, although machine PNT systems still fall short of the navigation intelligence endowed by the brain's spatial cognition, they possess distinctive strengths of their own. For instance, the absolute sensing accuracy and resolution of machine PNT systems exceed those of biological systems. Even a low-cost Inertial Measurement Unit (IMU) can readily achieve angular resolution on the order of $0.1°$. Moreover, machine PNT systems draw upon a far more diverse array of sensor modalities, such as LiDAR, millimeter-wave radar, and RF sensing, which are not available to biological senses. Furthermore, state-of-the-art atomic clocks developed by humans provide timing systems with extraordinary temporal resolution, drifting by only one second over hundreds of millions of years. This level of precision is unattainable by any biological brain. As the most advanced intelligent species on the planet, humans are uniquely capable of providing real-time global mapping and positioning services. In addition, humans possess the capability for textual description. Cartographic and maps of the environment can be regarded as extensions of this descriptive ability [7]. Distilled, codified, and embedded into algorithms and rules, these capabilities are transplanted into machine PNT systems, constituting a unique advantage that arises from human intellectual intervention.

Against this backdrop, we pose the following question: How can we ensure that unmanned systems are not only equipped with an intelligent "brain" driven by spatial cognition but also fully exploit the high-precision advantages of machine PNT—ultimately realizing abilities that emulate yet surpass the brain? This issue is imperative for brain-inspired PNT research and form the motivation of this paper. Therefore, this study seeks to dissect the technical characteristics of conventional machine PNT (exemplified by GNSS) alongside those of brain-inspired PNT, aiming to explore their conjunction pathways.

The main contributions are as follows.

1) This paper dissects the core characteristics and fundamental differences among various PNT systems (Section II).

2) It offers a reference framework for exploring how heterogeneous PNT paradigms can complement one another and be meaningfully integrated (Section III).

3) Building on the analyses of differences and Conjunction among these architectures, the paper presents perspectives and recommendations for the future development of brain-inspired PNT and, more broadly, universal PNT (Section IV).

*Outlines.* Section II is the differential dissection of diverse PNT systems. Section III is the exploration of heterogeneous PNT fusion paradigms. Section IV showcases discussions and perspectives. Section V is the conclusion.

## 2. Overview and Comparison of Different PNT Systems

This section dissects, at the levels of principle, roadmap, and hardware, the core characteristics and distinctions among biological brain PNT, conventional machine PNT, and emerging brain-inspired PNT architectures. By distilling the essential attributes of their respective characteristics, we aim to lay the foundation for subsequent heterogeneous PNT fusion while also offering a rapid cognitive outline for the broader PNT researchers.

### 2.1 Technical Principle

#### 2.1.1 Machine PNT System

Figure 1 illustrates the pipeline logic of the traditional machine PNT system.

***Spatial Metric Datum.*** The PNT capability of machine systems is grounded in human-defined, universally accepted spatial metrics, specifically coordinate system descriptions. For instance, humans have established various reference frames such as the inertial frame, the Earth-Centered Earth-Fixed (ECEF) frame, the local "East-North-Up (ENU)" frame, and etc., all of which are adopted as collective conventions [8]. Furthermore, given that each machine system has its own mechanical layout and sensor installation geometry, it is necessary to define body frames and individual sensor-centric frames.

***Spatial Sensing.*** Since perception is the prerequisite, whether the data fusion is loose or tight, the machine PNT system relies on high-precision sensors to sense the external world. Usually, GNSS can supply a unified spatiotemporal datum, allowing machine systems to describe the spatial relationships encountered during navigation and positioning. Nevertheless, GNSS is not omnipotent. Therefore, machine PNT systems often need to be equipped with other sensors to cope with GNSS denial and GNSS spoofing issues. These sensors leverage a diverse array of physical, chemical, and even bio-inspired mechanisms to detect spatial cues within the environment. Subsequently, these cues are translated into both intrinsic PNT information (e.g., attitude angles, velocity, time) and extrinsic PNT information (e.g., relative distance, relative direction) to meet the requirements of precise spatial description modeling.

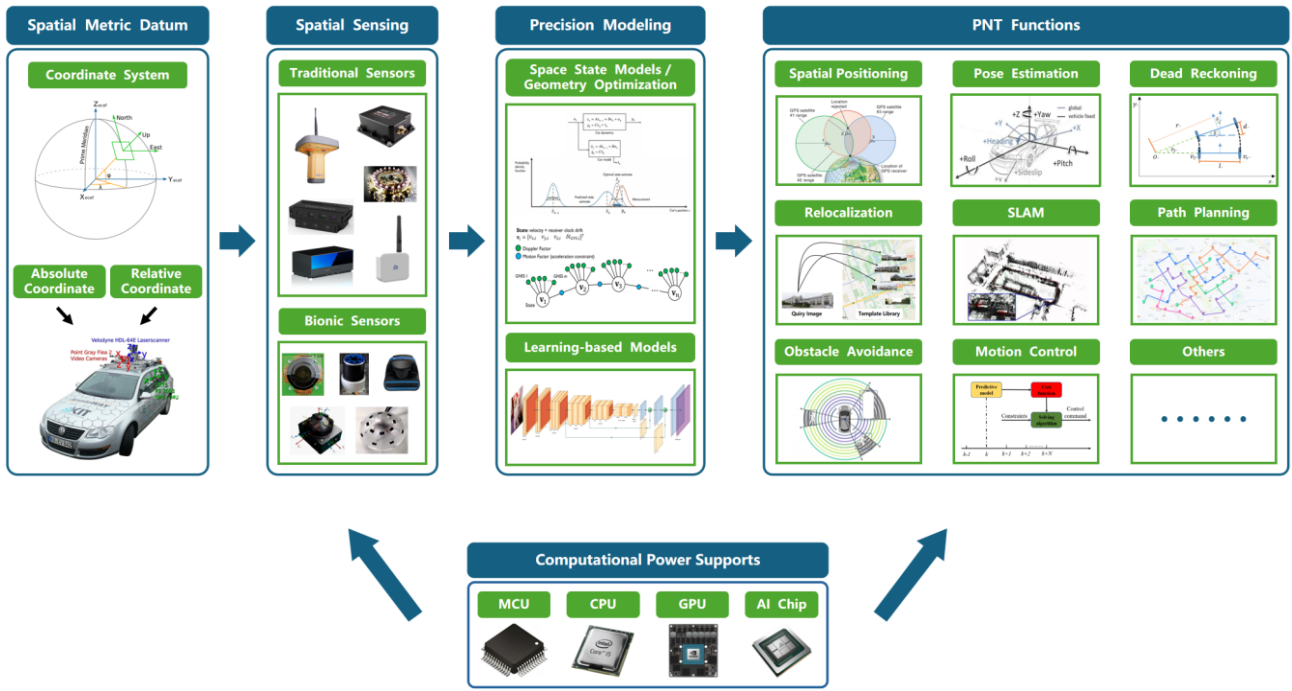

**Fig. 1** Characteristics of traditional high-precision PNT systems.

***Precision Modeling.*** Traditional PNT is rooted in a mathematical computational paradigm. When a machine system performs either egocentric or allocentric PNT tasks, it inevitably faces the need to unify and standardize disparate spatial metrics, i.e., perform coordinate transformations between various coordinate systems [9]. This presupposes that the spatial constraints between these frames are precisely known. Otherwise, the machine system will struggle to answer the two fundamental questions, "Where am I?" and "Where am I going?".

For example, traditional PNT systems, exemplified by GNSS, rely on human-engineered rules, such as satellite orbital models, error correction models, ranging intersection models, and so forth [10][11]. The entire workflow is governed by rigorous engineering standards and specifications, and hinges on precise computation and prescribed rules. The emphasis of this process prioritizes the accuracy and reliability of the computed result, striving to deliver precise PNT outputs via rigorous mathematical logic and meticulous engineering. The computational methods involved are all man-modeled numerically to suppress uncertainty in spatial modeling, such as space state models (e.g., Kalman filtering, particle filtering, etc.) and geometric or probabilistic optimization models (e.g., graph optimization, gradient- or residual-based optimization methods) [12]. However, the man-modeled numerical computation rules are not always reliable for the diagnosis of observation noise and modeling error uncertainty in PNT systems. To mitigate PNT errors, machine PNT systems also need to address the compensation of observation errors and the quantification of model uncertainty while constructing their spatial description models.

In recent years, numerous studies have integrated AI computation paradigms into PNT systems. Using data-driven methods to address issues like observation error compensation and uncertainty estimation, these studies have outperformed traditional numerical techniques. Undeniably, AI has made a positive, transformative contribution to the intellectualization of traditional PNT systems, particularly in tasks such as Simultaneous Localization and Mapping (SLAM), path planning, and motion control, etc. Yet, at their core, these gains remain within the realm of statistical learning. The representations automatically extracted from data typically capture only limited correlations between features/patterns and tasks (i.e., the maximum joint-probability distribution), yielding a biased, superficial depiction of complex real-world associations. Notably, since statistical models encode correlations rather than causation, they are prone to failure when data distributions change [13]. Because correlation does not equal causation, out-of-distribution generalization remains the Achilles' heel of AI paradigms [13][14].

*PNT Functions.* Based on the precise spatial modeling, machine systems' PNT functions can be implemented centered on the "perception-planning-control" framework. Their PNT capabilities at the perception and planning levels mainly include Dead Reckoning (DR), SLAM, path planning, obstacle avoidance, etc. Specifically, since mechanical carrier motion control mainly focuses on driving structures like motors and servos, this paper separately addresses the motion control capabilities. Common ideas include traditional Proportion-Integral-Differential (PID)-based control, fuzzy adaptive control, and Active Disturbance Rejection Control (ADRC), as well as learning-based (e.g., reinforcement learning) control approaches [15].

*Computational Power Supports.* Typically, conventional PNT systems frequently leverage Microcontroller Units (MCUs), Central Processing Units (CPUs), Graphics Processing Units (GPUs), or AI-dedicated chips to provide the requisite computational power supports, thereby facilitating the integrated implementation of diverse PNT capabilities. Moreover, in machine PNT systems, MCUs and CPUs are typically also tasked with communication and external command processing, etc. However, these aspects are not the focus here and thus will not be elaborated upon.

*Summary.* In summary, the high-precision performance of conventional machine PNT is predicated on human-unified spatial metrics, meticulous measurements, and exact spatial description and modeling. In essence, this is a "tool-oriented" application paradigm. When these foundations fail, the PNT capability of machine systems can fall into severe crisis. Although many researchers are actively leveraging AI-enabled, data-driven learning approaches to refine or even reshape human-designed spatial metric and description rules, these methods still grapple with prominent challenges, such as weak generalization and poor interpretability.

### 2.1.2 Brain PNT System

As common knowledge in neuroscience, cognitive maps are fundamental to spatial cognition [16]. The numerous external spatial cues in the physical world, combined with the internal states of organisms as they interact with and explore this world, constitute all the elements for building cognitive maps in the brain. These enable the brain to develop various PNT capabilities through its evolved neural circuits. Figure 2 illustrates the pipeline logic of the biological brain PNT system.

*Sensorium System.* Organisms rely on their sensoria, such as vision, hearing, smell, touch, and proprioception, to capture spatiotemporal cues from their internal state and the external world. Though less accurate and sensitive than man-made instruments, biological sensory systems can seamlessly convert spatial cues into physiological electrical signals without issues like downsampling, or analog-to-digital conversion. Moreover, through their sensorium system, an organism can effectively obtain spatiotemporal cues that support its own PNT capabilities, which are then processed collaboratively by different navigation cells. This embodied mode minimizes reliance on precise sensing methods and reduces energy consumption.

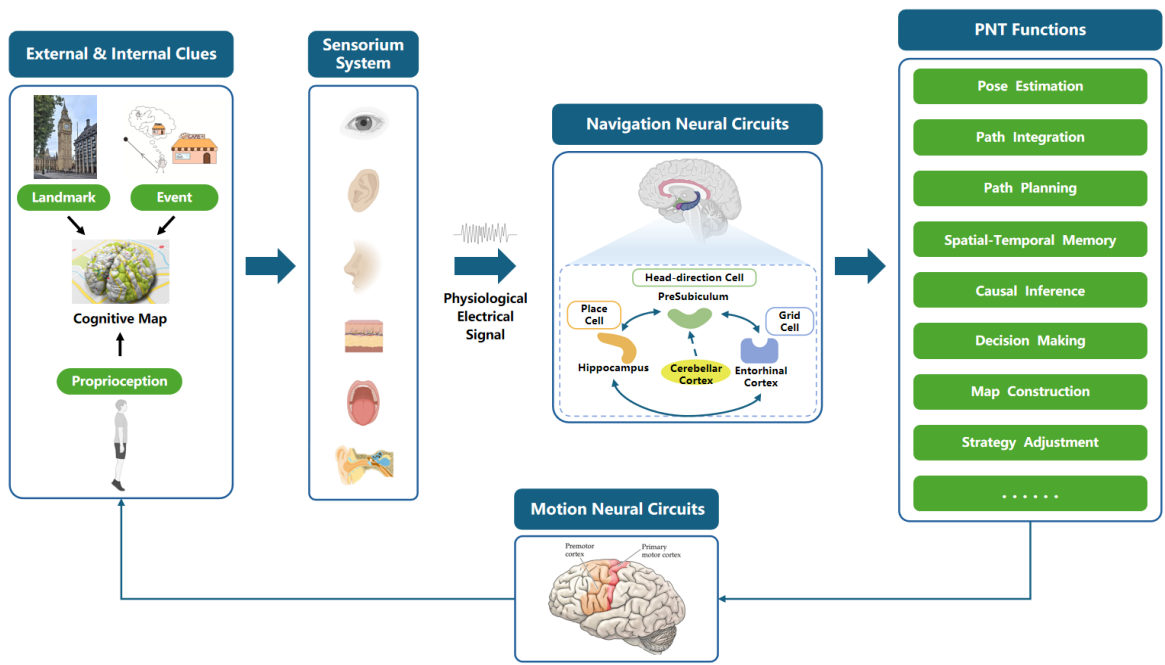

**Fig. 2** Characteristics of biological brain PNT systems.

*Navigation Neural Circuits.* Brain's navigation neural circuits operate through a fuzzy spatial representation without reliance on exact metrics or models, and all processing utilizes a unified neural signal format. Therefore, it can circumvent the loose-coupling vs tight-coupling issues encountered by machine PNT systems.

Per neuroscience, the hippocampus-entorhinal cortex system, rich in diverse navigation cells, is key to building cognitive maps and enabling spatial cognition and navigation in brains [7][16]. All spatial cues captured by the sensorium system are integrated within neural circuits that are populated by these specialized navigation-related neurons. For example, place cells, grid cells and head-direction cells, known as the "big three", perform as brain's GPS, coordinate system, and compass, respectively [6][17]. Moreover, the conjunction and collaboration of the "big three" with other navigation cells can endow the brain with capabilities for spatial perception, path integration, localization and mapping, and the formation of spatial memories [18]. For instance, grid cells and boundary cells can work together to assess environmental scale and depth from visual input [19]. The collaborative neural mechanisms between spatial-view cells, goal-vector cells, landmark cells, time cells, and speed cells are all profitable for spatial cognition capabilities [20]-[24].

After compiling internal and external spatial cues, the brain must also represent and transform between egocentric and allocentric spatial representations [20]. Without relying on numerical coordinate transformation rules, the brain's navigation neural circuits adaptively regulate the process based on environmental interactions [21]. This is enabled by the coordination of head-direction cells, spatial-view cells, and diverse boundary cells, etc. for egocentric and allocentric spatial coding [22][23]. Additionally, a unique class of egocentric neurons has been identified in the human brain, which, along with place cells and grid cells, forms the brain's GPS system [24]. These egocentric neurons encode the positions and orientations of external objects and the environment, establishing an internal "front/back, left/right" coordinate frame to support spatial recognition and goal-directed navigation. In addition, many animals, particularly social species, possess shared coordinate systems that describe local space. These systems are typically centered on nests, migratory destinations, habitual workplaces, or salient landmarks. However, these shared frames are not explicit computational constructs akin to human spatial metrics. Rather, they are long-term spatial experiences that evolve from the brain's autonomous utilization of cognitive maps and spatial memories.

*PNT Functions.* The brain's PNT capabilities, rooted in cognitive maps and spatial experience, span fundamental skills (e.g., pose estimation, localization and mapping, path integration, path planning, and etc.) and advanced abilities (e.g., spatiotemporal memory, causal reasoning, strategy optimization, and etc.). The fundamental PNT capabilities align with the overall objectives of "perception-planning-decision" pipeline. Moreover, the more advanced PNT capabilities may be crucial factors for the construction of spatial experience [7][25][26], enabling brain's generalization from limited experience to broader PNT tasks. This mirrors Jiao et al.'s [27] view: extracting information from perception, knowledge from information, experience from knowledge, and wisdom from experience. However, currently, there are still many mysteries in the basic scientific research on the brain's advanced spatial cognition capabilities that remain unsolved.

*Motion Neural Circuits.* Given the neuroanatomical distinctions of different functional neural circuits in the brain, this paper treats the motor neural circuit as an independent module rather than integrating it with the hippocampus-entorhinal cortex system. Following the completion of decision-making by the navigation neural circuits, the motor neural circuits must translate related neural signals into motion control commands for the biological organism. These brain regions typically include the premotor cortex, primary and secondary motor cortex [28][29].

*Summary.* In summary, the brain possesses no artificially defined, globally unified spatiotemporal datum and no numerical coordinate concepts such as world or polar frames. Instead, its PNT capability emerges from a complex neural network composed of diverse navigation-related cell types distributed across multiple brain regions. This network constructs a cognitive map and spatial experience of the external physical world, thereby equipping the brain with spatial cognition and navigation intelligence. During spatial cognition and navigation, the animal brain relies heavily on intricate neural activities that are closely associated with mind, emotion, and consciousness. These activities range from innate approach-avoidance behaviors to causal inferences that maximize benefits, all of which are governed by elaborate reward-and-punishment neural circuits. In essence, the core competence of the biological brain PNT system is an integrated cycle of perception, cognition, reasoning, decision-making, and control, representing a mentalization-driven operating mode.

### 2.1.3 Brain-inspired PNT System

Figure 3 illustrates the pipeline logic of nowadays brain-inspired PNT. Although early efforts date back two decades, (e.g., Milford et al.'s RatSLAM prototype in 2004 [30] and Nobel laureate Edelman team's 2007 Darwin robot that emulated hippocampal place cells for navigation [31]) progress of brain-inspired PNT remained sparse for many years. Only recently has this frontier attracted intense attention, with a steady stream of original advances emerging.

*Spatial Metric Datum.* Because the brain-inspired PNT systems are man-made engineered works, they cannot escape the need for coordinate-based spatial metrics. Current neuroscience has yet to fully elucidate how the brain performs unified coordinate transformations between egocentric and allocentric frames. Consequently, brain-inspired PNT at its current stage cannot realize a unified framework based purely on neural-circuit emulation and must continue to employ conventional PNT techniques to describe both absolute and relative coordinates. Moreover, the coordinate transformation issues caused by the installation methods of different sensors are also inevitable for brain-inspired PNT systems. These issues can be simply and efficiently resolved using traditional numerical computation methods. Therefore, they are unlikely to be eliminated by various man-made PNT systems.

*Spatial Sensing.* Like traditional machine PNT systems, current brain-inspired PNT systems also rely on high-precision observations to construct spatial description models, remaining within the realm of precise measurement. While maintaining compatibility with traditional and bionic sensors, brain-inspired PNT also supports the integration of neuromorphic sensors, which are characterized by high dynamics, ultra-low power consumption, and event-driven operation [32].

*Spatial Modeling.* Current brain-inspired PNT still relies on numerical computation to adhere to human-defined spatial metrics for describing spatial coordinates, align with traditional machine PNT. Despite brain-inspired PNT is still in its infancy, lacking a settled paradigm, yet it is already clear that the brain's navigational mechanisms provide a new theoretical lens for spatial modeling. Unlike traditional machine PNT, guided by spatial cognition principles, researchers are replacing hand-crafted spatial modeling with brain-inspired neurodynamic models. This method can bypass artificially designed deterministic models to handle uncertainties in spatial modeling [33][34], even with limited observation precision. While less accurate than traditional numerical methods, its focus is on replicating the brain's spatial cognition through neurodynamic models, rather than achieving the absolute precision sought by conventional PNT systems.

The mainstream brain-inspired spatial modeling methods currently include Continuous Attractor Neural Networks (CANNs), Spiking Neural Networks (SNNs), and Deep Reinforcement Learning (DRL), etc. These models with neurodynamic simulation characteristics, all share the commonality of mimicking brain mechanisms. They draw inspiration from the brain navigation mechanisms, implementing PNT functions by simulating brain's neural activities at neuron-level, circuit-level or system-level [35]. However, not all these methods explicitly involve optimization processes like numerical calculation and data-driven computation paradigms [36]-[38].

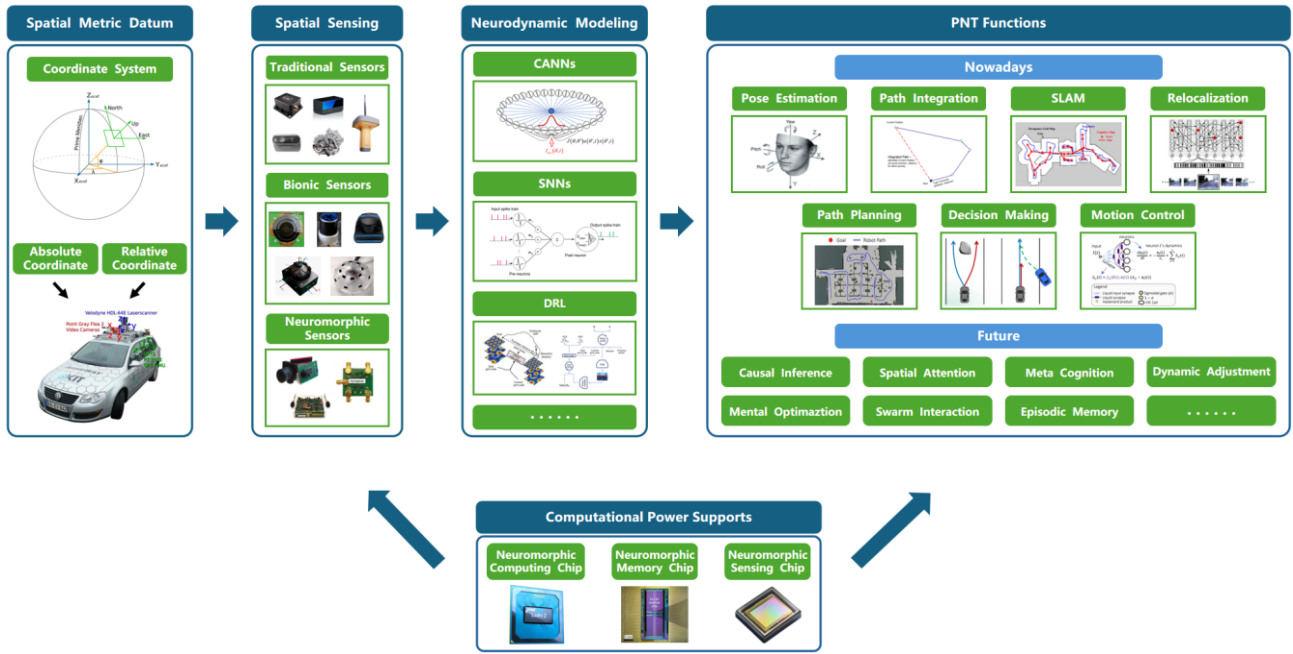

**Fig. 3** Characteristics of current brain-inspired PNT systems.

***PNT Functions.*** Currently, brain-inspired PNT is in an exploratory phase, with diverse technical ideas and entry points, ranging from emulating navigation cell patterns to studying how connectome patterns shape spatial cognition. The field focuses more on drawing innovative inspiration from brain neural mechanisms, aiming to break through conventional engineering constraints and explore new pathways for navigational functionality. However, the technical roadmap of nascent brain-inspired PNT largely follows the traditional "perception-planning-control" framework. Thus, its capabilities are mainly seen in pose estimation, SLAM, path integration, path planning, decision-making, and control, etc. Only a few studies, such as [39]-[41], have initiated explorations into more advanced spatial cognition abilities like complex memory, causal inference, and meta-cognition. We believe these higher-level spatial cognition abilities will be the focus of future bio-inspired PNT development.

For example, at the perception level, RatSLAM and its variants focus on using CANNs to emulate the brain's path integration capability, integrating spatial cues and mapping spatial experience into a topological experience map. Although this approach yields lower mapping accuracy than traditional methods like factor-graph optimization in iSAM [43], and bundle adjustment in ORB-SLAM [44], it better reflects how organisms naturally perceive space. Apart from the reconstruction of the backend optimization, RatSLAM and its evolutions, such as NeuroSLAM [45], do not change the existing SLAM framework. ***Note:*** We have open-sourced the Python version of NeuroSLAM (URL: https://github.com/BINUCOE/OpenNeuroSLAM). It has been benchmarked against the open-source MATLAB version [45] to aid in understanding the algorithm at the code level.

At the planning level, Steffen et al. [46] developed a 3D path planning method inspired by hippocampal place cells and synaptic plasticity. Instead of using greedy algorithms or reinforcement learning, their method treats synaptic weights as a vector field to generate a trajectory from start to goal, producing a biologically plausible path, though not always the shortest. Liu et al. [47] reported a brain-inspired decision-making method, even outperforming conventional algorithms in planning efficiency. We conjecture that, compared with RatSLAM's dense storage of spatial experience, the gains reported stem from an aggressive sparsification of the spatial experience.

Likewise, progress has been made in brain-inspired control. For instance, Paredes-Vallés et al. [48] described a fully neuromorphic visual-control pipeline for UAVs. While the SNN controller's accuracy is lower than that of PID controllers, this approach nonetheless represents a valuable alternative.

***Computational Power Supports.*** Unlike the von Neumann architecture-based processors widely used in traditional machine PNT systems, neuromorphic chips, inspired by neuroscience, has empowered brain-inspired PNT systems with novel advantages [49]. These chips, emulating the brain's integrated memory-computation mode, offer outstanding advantages in terms of energy efficiency, latency, and power consumption. Details see Section 2.2.

***Summary.*** In summary, align with traditional machine PNT, nowadays brain-inspired PNT also depends on human-unified spatial metrics and precise measurements. However, unlike traditional machine PNT, brain-inspired PNT shifts the principles of spatial description and modeling. It simulates the brain's spatial cognition mechanisms via brain-inspired neural networks, instead of relying on artificially designed numerical computation rules. Moreover, compared with traditional machine PNT systems, brain-inspired PNT systems introduce the increment of neuromorphic hardware while being compatible with traditional sensors and computing units. Thus, the current gap between brain-inspired PNT and traditional machine PNT lies in hardware implementation and computational methods. However, at present, brain-inspired PNT is still confined to the instrumental pipeline of "perception-planning-control." There is still a huge gap between the current achievable brain-inspired PNT capabilities and the complete simulation of the brain's mentalization-driven navigation intelligence.

## 2.2 Hardware

*Sensors.* Hardware is a critical pillar of any PNT architecture. Both traditional and brain-inspired PNT rely on diverse sensors to endow the system with machine perception. Traditional sensors, including bionic ones, do not utilize neuromorphic computing principles. They transduce physical phenomena (acoustic, optical, mechanical, chemical, etc.) into analog or digital signals. Their dynamic range and latency are dictated by fabrication processes and sampling rates, and they generally consume considerable power. By contrast, the rapidly emerging neuromorphic sensors impart PNT systems with entirely new perceptual capabilities. Operating on neuromorphic computing principles, they work in an event-driven fashion and represent sparse spatiotemporal events with spike trains, offering high dynamic range and ultra-low power consumption [49]. Biological senses, in turn, process information via unified neuro-electrophysiological signals, achieving extremely low latency and power draw. Yet they fall short of artificial devices in the richness, configurability, and communication bandwidth of their sensing modalities. Table 1 summarizes the characteristics of current PNT sensors alongside those of biological perception.

**Table 1** Traditional multimodal sensors vs. brain-inspired sensors

| | | Traditional Sensors (including bionic sensors) | Brain-inspired Neuromorphic Sensors | Biological Senses |
|---|---|---|---|---|
| **Perception Modality** | Vision | Mono/Stereo, Depth, Panoramic cameras, Compound eyes [50], Lidar, etc. | Event cameras [51] | Eye |
| | Kinesthetic | IMU, Bionic campass [52], etc. | — | Vestibular system |
| | Auditory | Conventional microphones, Biomimetic auditory sensors [53], etc. | Dynamic auditory sensors [54] | Ear |
| | Olfactory | Electrochemical/optical gas sensors [55], Electronic nose [56], Bionic olfactory sensors [57], etc. | Neuromorphic olfactory chip [58], Neuromorphic E-nose [59] | Nose |
| | Tactile | Capacitive / Piezoelectric tactile sensors [60], E-skin [61], etc. | Neuromorphic tactile sensors [62] | Skin |
| | Radio | Wi-Fi / UWB / LoRa / Bluetooth, etc. | Neuromorphic radar [63], Neuromorphic radio frequence sensors [64] | — |
| | Acoustic | Ultrasonic sensors, Sonar, Bionic sonar [65], etc. | — | Vocal tract |
| | Geomagnetic | Magnetometers, Bionic magnetic sensors [66], etc. | — | Magnetoreceptors |
| | Communication | 4G/5G, etc. | — | Language, Pheromones |
| **Feature Comparison** | Signal Processing | Analog → ADC → Digital | Pulse (event) signals | Electrophysiological signals |
| | Data Representation | High-redundancy raw data | Sparse spatiotemporal event data | Neural activities |
| | Power Consumption | Typically ≥ mW | Typically μW | μW |
| | Latency | High (sampling-rate-limited) | Low | Low |
| | Dynamic Range | Limited | High | High |

*Processors.* Beyond sensors, chips and processors are the vehicles that bring PNT technologies into real-world deployment. Under the von Neumann architecture, we have witnessed the rise and widespread adoption of high-performance CPUs/GPUs and AI-dedicated chips. Even as manufacturing processes approach industrial limits and Moore's Law falters, Graphcore's IPU has achieved the astonishing feat of integrating more than 23.6 billion transistors on a single die [67]. Yet the von Neumann paradigm's separation of memory and computation creates an inherent trade-off between power efficiency, latency optimization, and raw computational growth. Neuromorphic computing, inspired by neuroscience, emulates the brain's spatiotemporal processing and in-memory-compute paradigm, giving birth to a new generation of brain-inspired chips/processors. Compared with their von Neumann counterparts, neuromorphic devices use analog neurons and synapses as primitive computing elements, offering markedly lower power and latency.

For instance, Tang et al. [68] implemented an SNN-based SLAM algorithm on Intel's Loihi chip and demonstrated comparable accuracy to the widely used GMapping method. Their neuromorphic SLAM consumed 100× less dynamic power on Loihi than GMapping running on a conventional CPU. Similarly, Yoon et al. [69] introduced NeuroSLAM, a neuromorphic SLAM IC that, via mixed-signal oscillators, achieves ultra-low power (< 24 mW) and high energy-efficiency (> 7 TOPS/W) for edge SLAM, while delivering real-time performance (> 100 fps) and rapid path integration and loop closure detection, ideal for power-constrained edge robots. Recent neuromorphic chips even feature on-chip learning. Examples like Tianjic and SpiNNaker2 further support hybrid heterogeneous deployment of both AI and brain-inspired paradigms [70]-[72]. Table 2 summarizes the characteristics of chips and processors across these different architectures.

**Table 2** Traditional processors vs. brain-inspired processors

|  | **Architecture Design** | **Memory** | **Computing Elements** | **Learning Capability** |
|---|---|---|---|---|
| **Traditional Processor** | von Neumann-based | Memory and computation separated | Transistors and logic circuits | Software-driven; external training required |
| **Neuromorphic Processor** | Neuromorphic-computing-based | In-memory / compute-in-memory | Analog neurons and synapses | Synaptic plasticity (e.g., STDP) implemented in hardware; weights adapt dynamically |

However, integrating heterogeneous hardware into PNT systems today faces a practical obstacle. Because of engineering issues such as data-protocol compatibility, neuromorphic sensors and traditional sensors still require von Neumann modules as intermediaries before they can be merged into a larger PNT stack. These transition boards reconcile communication between disparate subsystems. As a result, most architecturally complex unmanned platforms can only swap in neuromorphic chips at the compute stage, replacing the von Neumann CPU/GPU to gain algorithm-level brain-inspired optimizations, while leaving the rest of the perception-to-control chain in its conventional form. The drawback is that an end-to-end neuromorphic hardware pipeline remains elusive. That said, researchers have reported near-complete neuromorphic scheme for small-scale drones whose system complexity is low, such as [48].

### 2.3 Summary

The distinctions of different PNT systems are summarized in Table 3. Whether it is a traditional PNT system or a brain-inspired PNT system, the prerequisite for realizing PNT functions is to build a compatible and usable spatial description model (i.e., spatial modeling). Moreover, as analyzed in section 2.1, both rely on human-defined unified spatial metric datum. However, they differ in their focus on spatial description and modeling, as well as in the corresponding implementation techniques.

Overall, traditional machine PNT relies on human-defined spatial metrics and precise measurements, forming a "tool-oriented" paradigm that faces crises when these foundations fail, despite AI-driven efforts to refine them. In contrast, the brain's PNT capability emerges from a complex neural network across multiple regions, constructing cognitive maps without artificial spatiotemporal datums or numerical coordinates. This biological system integrates perception, cognition, reasoning, decision-making, and control, driven by mental, emotional, and reward-related neural activities.

**Table 3** Comparison of underlying principles across different PNT systems

|  | Machine PNT | Brain PNT | Brain-inspired PNT |
|---|---|---|---|
| Spatial Metric Datum | Human-defined unification | Spatial experience | Human-defined unification |
| Spatial Sensing Mode | Precise observation | Neural expression | Precise observation (including neuromorphic sensing) |
| Spatial Description | Precision modeling with human-engineered rules | Cognitive maps with spatiotemperal memory | Neurodynamic modeling guided by brain-inspired spatial cognition rules |
| Computational Paradigm | Numerical or learning-based paradigms | Biological neural networks and systems | Neurodynamic simulation paradigms |
| Function Realization | Perception-Planning-Control (tool-oriented pattern, emphasizing high-precision computation) | Perception-Cognition-Reasoning-Decision-Control (mentalization-driven pattern, emphasizing spatial cognition & mental optimization) | Perception-Cognition-Control (nowadays, tool-oriented pattern) Perception-Cognition-Reasoning-Decision-Control (future, cognition-driven pattern) |
| Hardware | von Neumann-based sensors (including bionic sensors) and processors | Biological sensorium system and the whole brain | Both von Neumann-based and neuromorphic sensors and processors |
| Commonality | All require unified egocentric/allocentric spatial metric transformations, and an usable spatial description model |||

Brain-inspired PNT, while still using human-unified metrics and precise measurements, shifts to simulating brain mechanisms through neurodynamic models, differing from traditional machine PNT in hardware and computation principle. However, its capabilities remain rudimentary and shallow, falling short of fully simulating the brain's mentalization-driven navigation intelligence. Some researchers are trying to integrate higher-order spatial cognition, like memory and reasoning, into PNT technologies. However, these efforts are advancing in isolation, and several technical barriers remain. By learning from neuroscience and developing brain-inspired computational models, the future evolution of brain-inspired PNT should break from traditional architectures and adopt a new framework of "perception-cognition-reasoning-decision-control." This shift will transform PNT from a "tool-oriented" to a "cognition-driven" paradigm.

## 3. Framework of the Fusion Strategies between Brain-Inspired and Traditional PNT

In essence, the divide between traditional PNT and brain-inspired PNT is the distinction between engineering precision and brain-inspired intelligence, while their shared reliance on spatial description models provides a bridge for technical dialogue. As technology evolves, the two should move from "competitive divergence" to "complementary synergy," jointly forging a more ubiquitous, intelligent, reliable, and robust PNT architecture to meet the diverse and complex demands of the future. Therefore, this section explores fusion pathway between traditional PNT and brain-inspired PNT, aiming to provide a reliable technical path for PNT systems operating in GNSS-denied environments so they can exhibit greater resilience in complex scenarios. Figure 4 presents an architecture diagram of the proposed fusion framework. It comprises two parallel pipelines: traditional PNT and brain-inspired PNT.

First, it involves the integration and coupling of heterogeneous sensory at the observation level. Neuromorphic sensors are utilized to counter challenges such as low light, motion blur, and feature degradation, while simultaneously leveraging the comprehensive modality coverage of traditional sensing methods and their advantages in perceiving texture information. This facilitates more reliable and robust capture of environmental perception data to support the fusion of heterogeneous data at the observation level.

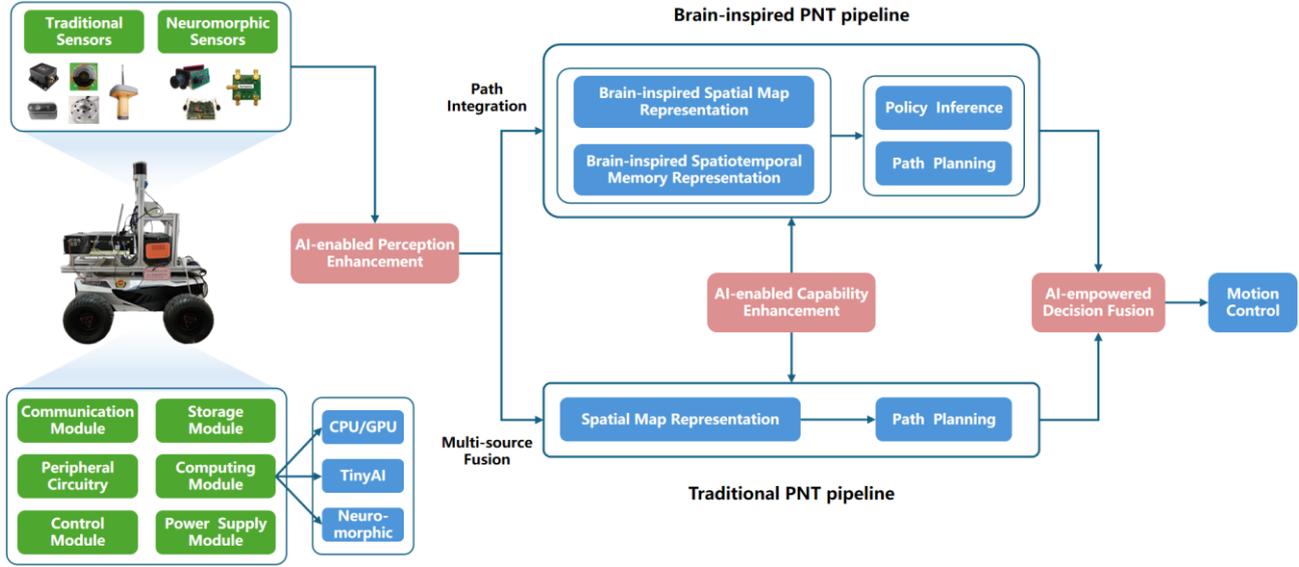

**Fig. 4** The architecture diagram of the proposed fusion pipeline.

Second, at the capability level, after heterogeneous perception data undergoes preprocessing such as time synchronization, certain modalities can be enhanced using AI augmentation methods. For example, it can be used to perform error compensation on UWB ranging observations [73], automatically extract point/line texture features of visual odometry [74][75], and suppress observation noise of IMUs [76], among other tasks. The AI-enhanced perception modality not only reduces uncertainties in spatial description models caused by observation errors for traditional mathematical PNT pipelines but also improves the accuracy of spatial cues in the construction of brain-inspired spatial description models, which is beneficial for enhancing the effectiveness of spatial experience decoding. Moreover, in processes such as spatial mapping, track estimation, and path planning, some mature and reliable AI methods can exert their advantages in subtasks such as relocalization, semantic segmentation, dynamic parameter regulation, and motion control adjustment.

Third, at the decision level, high-precision PNT technologies, exemplified by GNSS, rely on robust least squares and space state models (e.g., Kalman filtering) to process signals from satellites and other sources. Based on this, decision-making can be enabled through perception-based planning to drive the PNT system's motion control. While AI methods have been integrated into the traditional PNT pipeline to enhance perception and planning capabilities, the underlying logic mentioned above remains unchanged. However, no prior research has addressed how to decouple from the traditional PNT pipeline and integrate it with the decision-making capabilities of the parallel brain-inspired PNT pipeline shown in Figure 4. Moreover, we believe that the significance of this issue will become increasingly evident as the field of brain-inspired PNT progresses. Using AI methods such as Mixture-of-Experts (MoE) strategy [77] or voting mechanism in ensemble learning to derive reliable decision fusion strategy is the current consideration in this paper. Moreover, we believe that this issue deserves continuous in-depth investigation, especially when the paradigm of brain-inspired PNT transcends the conventional "perception-planning-control" framework.

Finally, hardware-level fusion must address both sensor and compute processor integration. In principle, all algorithms in either the traditional or brain-inspired PNT path can be deployed on state-of-the-art hybrid neuromorphic chips (e.g., Tianjic, SpiNNaker). However, due to the current limitations of underdeveloped toolchain and inadequate ecological support, users often encounter numerous engineering issues when deploying heterogeneous computing models. Therefore, edge AI chips and some dedicated AI acceleration hardware may serve as alternatives when full neuromorphic deployment fails. Additionally, issues such as architectural compatibility and data protocol alignment between different hardware modules for communication, control, storage, and computing need to be addressed from an engineering perspective during hardware system integration.

### 3.1 Observation-Level Integration

At the observation level, a machine PNT system's environmental awareness originates from its suite of multimodal sensors and signal-sensitive devices. Traditional multimodal sensors already cover a very broad range of sensing modalities, whereas neuromorphic sensors have so far achieved breakthroughs mainly in vision, hearing, olfaction, and taste, etc. Although neuromorphic perception injects new vitality into PNT systems, it cannot solve every practical challenge on its own. Consequently, many researchers now combine conventional and neuromorphic sensors to attain more powerful general-purpose perceptual capabilities.

For example, NeuroGPR [78] is a universal place recognition system that fuses traditional multi-source perception with neuromorphic vision. RAMP-VO, introduced by Pellerito et al. [79], is a novel visual odometry technique that blends conventional imagery with event camera data. Others have built Visual-Inertial Odometry (VIO) systems that fuse RGB-event hybrid visual features with IMU measurements to counteract weak lighting, motion blur, and feature degradation, as exemplified by TEVIO [80]. We contend that wheel odometry, UWB, and other sensing modalities could likewise be coupled at the observation level with heterogeneous vision sensors and IMUs to further enhance robustness.

Beyond the above, the United State (U.S.) Defense Advanced Research Projects Agency (DARPA) laid down an important conceptual marker in 2010 with its All-Source Positioning and Navigation (ASPN) initiative. One of the long-term goals of brain-inspired PNT is to endow machine systems with the same kind of resilience that ASPN promises for ubiquitous navigation. Today, however, brain-inspired PNT is only beginning the transition from single-source to multi-source sensing, and it still falls short of true brain-inspired ASPN.

The gap is most evident in the weak coupling between current brain-inspired PNT research and many mature sensing modalities. Fortunately, a handful of studies have already explored brain-inspired schemes that leverage more general multi-source information. Milford's group, for instance, has reported integrating Wi-Fi and barometric data using place cell and other navigation cell models [81]. Liu et al. [82] introduced MM-NeuroPOS, a brain-inspired localization scheme that fuses MEMS sensors with terrestrial digital broadcast signals. Yang et al. [83] constructed a path integration model composed of multiscale navigation cells using CANNs, which can achieve better fusion of GNSS and IMU information than traditional filtering-based methods. We therefore contend that, alongside heterogeneous sensor fusion at the observation level, guiding brain-inspired PNT from multi-source toward all-source sensing will be a key direction for future development. However, there are some potential challenges. For instance, the integration of heterogeneous observation from traditional and neuromorphic sensors faces challenges in time synchronization and compatibility issues due to the inconsistent data protocols, data interfaces, and data formats. Furthermore, the fusion of heterogeneous multimodal sensors, accompanied by an increase in observation data volume, will place pressure on hardware storage resources.

### 3.2 Capability-Level Integration

At the capability level, since brain-inspired PNT is still in the exploratory phase, under the traditional PNT framework of "perception-planning-control", most researchers currently choose to start from the perception level and have achieved some breakthroughs in SLAM. Through technical and code-level analyses of several well-known existing brain-inspired navigation schemes, such as RatSLAM and NeuroSLAM, it can be observed that although current brain-inspired PNT research has introduced novel computational paradigms with neurodynamic properties, there are several issues.

For instance, the brain-inspired, neurodynamic models cannot yet cover every component of a PNT system. At present, these methods are most applicable to information fusion, path integration, and spatiotemporal memory tasks. In NeuroSLAM, for example, only the backend of the conventional SLAM pipeline is re-implemented in a brain-inspired fashion. By drawing on the mechanisms of brain's implementation of path integration, a path integration model constructed using CANN transforms the uncertainty optimization problems into encoding-decoding problems of spatial experience. The entire technical route is not entirely realized using neurodynamic models, other components besides the backend still employ traditional methods relying on numerical models. This implies that other components, such as visual template matching for loop closure detection, can incorporate AI-enabled methods for targeted optimization and enhancement of capabilities, as seen in [84][85].

Furthermore, under the framework of encoding integration of spatial cues and decoding of spatial experience, the more accurate the encoded information, the more reliable the decoded experience should be. Therefore, enhancing perceptual modalities is beneficial because more precise measurements help reduce uncertainties in spatial description models. Therefore, deep learning techniques, which excel at pattern enhancement tasks, and even Tiny Machine Learning (TinyML) [86] techniques, can find their niche.

For another example, the self-organizing and adaptive interaction mechanisms among navigation cells in the brain's navigation circuits are still largely unknown. As a result, current brain-inspired PNT studies rely almost exclusively on static simulation approaches. This limits their ability to fully replicate the brain's spatial cognition intelligence. Take time cells as an example, researchers have so far only abstracted and simplified them. References [78] and [87], for instance, use a CANN to encode timing cues from an IMU or map continuous event-camera data into multiple bins via SNNs to mimic time cell functionality. Yet key properties of time cells, such as multiscale dynamics, periodicity, and scalability, are absent from these simplified models. Neuroscience has revealed that time cells cooperate with grid and place cells to encode episodic memories and event information [88][89], but we still lack non-static models capable of capturing such interactions. Similarly, dynamic modulation models for cross-dimensional collaborative interactions between navigation cells (e.g., 2D and 3D, 3D and 3D, 1D and 3D interactions), as well as cross-scale interaction modulation models among navigation cells with multiscale characteristics, have not been reported in current research.

Consequently, we argue that the current use of static simulation methods in brain-inspired PNT research, which lack dynamic modulation capabilities, makes it challenging to replicate the self-organization and adaptive regulation abilities of the brain's navigation neural circuits. Currently, the emerging dynamic neural network [90][91] technologies in the field of deep learning may hold promise as a solution to this impasse. In fact, research by Xu et al. [92] can serve as evidence. They analyzed the dynamic properties of attractor networks constructed based on Recurrent Neural Networks (RNNs) and found that the network exhibits different types of attractor properties under varying learning objectives. Predicting positions enables the network to form continuous attractor states, facilitating navigation tasks based on path integration. Predicting landmarks leads the network to generate discrete attractor states, conducive to landmark-based navigation tasks. They proposed a modular RNN model that combines two pre-trained RNN-based attractor models, enabling flexible switching between different navigation modes driven by reinforcement learning.

Beyond above, current brain-inspired neural networks remain rudimentary and limited in reasoning and decision-making tasks. Large language Models (LLMs), although not inherently adept at spatial cognition, possess vast repositories of expert knowledge that confer unique strengths in reasoning and decision-making. This characteristic has attracted many researchers to conduct research on embodied navigation that integrates perception and control based on LLMs [93]-[95]. Beyond this, higher-order capabilities emblematic of the brain, such as metacognition and meta-learning [96], transfer learning [97], lifelong learning [98], and causal learning [99], are pivotal for steering brain-inspired PNT from its current "tool-oriented" stage toward genuine "cognition-driven". However, current research on these topics in the field of brain-inspired intelligence is still in its infancy, though they have undergone a certain period of development and application in the field of deep learning.

Therefore, we advocate that, given the current primary stage of development of brain-inspired PNT, a "synergy of strengths" principle should be adopted, allowing numerical models, AI methods, and brain-inspired approaches to fully leverage their respective advantages within the architecture of brain-inspired PNT systems. This would enhance the usability and practicality of brain-inspired PNT systems, transcending mere theoretical verification. However, currently, integration in the capability-level lacks a unified architecture or theory for heterogeneous computing paradigms. It is recommended to continuously and incrementally replace and restructure various computing modules within the existing PNT framework, iteratively exploring the optimal combination.

### 3.3 Decision-Level Integration

At present, brain-inspired PNT does not outperform mature traditional PNT in absolute accuracy. Yet, thanks to its brain-simulated PNT mechanisms, it can achieve highly robust performance in complex environments. Moreover, when deployed on neuromorphic hardware, it surpasses traditional PNT in real-time capability and energy efficiency. For these reasons, we argue that a practical unmanned platform operating in demanding environments should carry both a traditional PNT system and a brain-inspired PNT system, with their outputs fused at the decision level to trade off availability, accuracy, and continuity.

However, this raises the question of how to utilize the outputs from different PNT pipelines at the decision stage in a reasonable manner. Even a simple weighted average requires a principled way to quantify the contribution of each pipeline. For example, in structured or information-rich settings the traditional system could receive a larger weight to safeguard precision, whereas in degraded-perception, no-prior, non-cooperative scenarios the brain-inspired system could be given precedence. Yet turning this intuition into an operational weighting scheme is non-trivial.

Currently, there is a significant research gap in the theoretical foundations and methodologies for effectively fusing traditional and brain-inspired PNT at the decision level. Specifically, for a PNT system equipped with both traditional and brain-inspired PNT abilities, questions arise: Should brain-inspired PNT run in parallel as an auxiliary mode to traditional PNT during tasks [100]? If so, how should their decision outputs be weighted and averaged? In this preliminary exploration, we believe that strategies like MoE and voting mechanism are worth trying, except relying on empirical tuning, for the following reasons.

Here, this paper draws on the theory of ensemble learning, which combines multiple branch sub-models in the hope of obtaining a more powerful and comprehensive model overall [101]. The underlying idea is that even if one of the weaker sub-branches produces an unreliable output, a meta-learner can weigh the outputs of the other branch models and correct the overall uncertainty of the ensemble learning model. This process coincides with the heterogeneous integration of traditional PNT and brain-inspired PNT at the decision-making level. When traditional PNT and brain-inspired PNT pipelines work in parallel, they can be regarded as two different branch sub-models in an ensemble learning framework. At this time, it is necessary to train a reliable meta-learner to achieve a balance between the two. Typically, the meta-learner in ensemble learning can be implemented using a voting model. In addition to this, given that the MoE strategy can achieve adaptive weighting by combining the outputs of multiple sub-models (experts) through weighted summation and dynamic weights adjustment based on different input characteristics, we believe it can serve as a meta-learner to meet our needs.

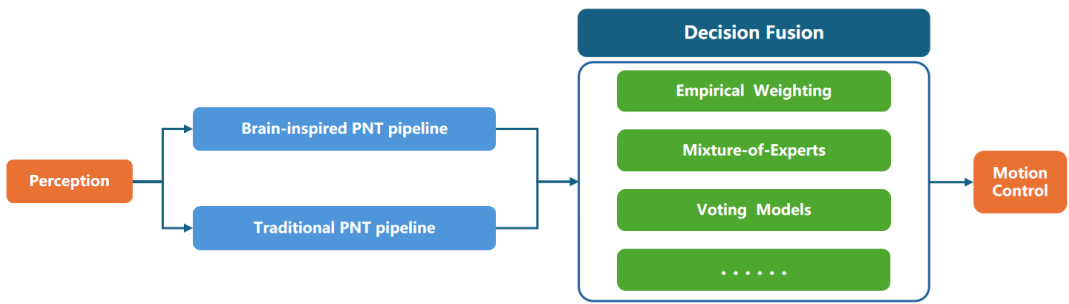

**Fig. 5** The roadmap for the heterogeneous integration of traditional and brain-inspired PNT pipelines at the decision-making level.

Figure 5 is the roadmap for the heterogeneous integration of traditional and brain-inspired PNT pipelines at the decision-making level proposed in this paper. However, some potential challenges need to be stated. For example, there is a scarcity of standardized testing platforms and methodologies specifically designed to both train these models and reliably validate their performance. Moreover, during the execution of PNT tasks, machine systems often face more refined local dynamic planning, obstacle avoidance, and even multi-agent interaction. Therefore, the integration of traditional and brain-inspired PNT at the decision level still needs to be further explored. In addition, security vulnerabilities and data privacy concerns inherent in the decision fusion process need to be comprehensively addressed.

### 3.4 Hardware-Level Integration

End-to-end neuromorphic hardware stacks are highly promising [102]. Yet, in most current machine PNT systems, heterogeneous sensors still follow distinct data protocols, and the functional modules for communication, control, storage, and computation remain physically discrete. Take the NeuroGPR system, it still relies on conventional computing processors to manage data processing, communication, and actuation for both neuromorphic and traditional sensors. Moreover, as previously stated, neuromorphic sensors have thus far only achieved breakthroughs in a limited number of perception modalities, and their capability for comprehensive environmental perception lags behind that of existing sensors. This implies that, at present, neither from the standpoint of sensor configuration nor from that of chip and processor configuration, are we close to the aim of realizing a fully neuromorphic machine PNT system.

Although some researchers have endeavored to configure small-scale machine PNT systems as comprehensively as possible using neuromorphic hardware solutions, such as [48]. However, this is attributable to the unique advantage conferred by the simplicity of the structure of small-scale machine systems and is not applicable to more extensive and general non-small-scale machine PNT systems. Therefore, this paper posits that employing heterogeneous sensor configurations to complement and enhance the perception capabilities of PNT systems is promising. Additionally, judiciously designing the configuration of traditional processors, AI processors, and neuromorphic processors with respect to computing, communication, storage, and control is crucial. Together, their conjunction represents the optimal strategy for the hardware-level integration of heterogeneous PNT systems at the current stage.

However, given that brain-inspired PNT is currently in its nascent exploratory phase, research on its heterogeneous integration with mature traditional PNT pipelines is even rarer. Therefore, the methodology for heterogeneous hardware integration oriented towards practical applications is not yet mature. Not only that, but also when considering heterogeneous integration at the hardware level, we are also confronted with a number of engineering problems that urgently need to be resolved. For example, the ecosystem of neuromorphic hardware lacks maturity, with insufficient support from comprehensive development, debugging, and deployment toolchains. There are some bottlenecks in achieving efficient communication between heterogeneous hardware components, and the implementation of precise hardware-level time synchronization. Moreover, when facing constraints inherent to specific application scenarios (e.g., on-device payload limitations, power budgets), it is often difficult to balance the energy optimization of the hardware system with its computational performance requirements.

Furthermore, the choice of data processing architecture is critical. Compared to the traditional approach where all sensor data is uniformly aggregated to a central processor, distributed processing mode may present a superior solution for many practical scenarios. This advantage stems from its low latency, reduced bandwidth dependency, and potential for enabling local real-time decision-making. Based on the above, Figure 6 is drawn in this paper as a reference for the hardware-level integration of a heterogeneous PNT system.

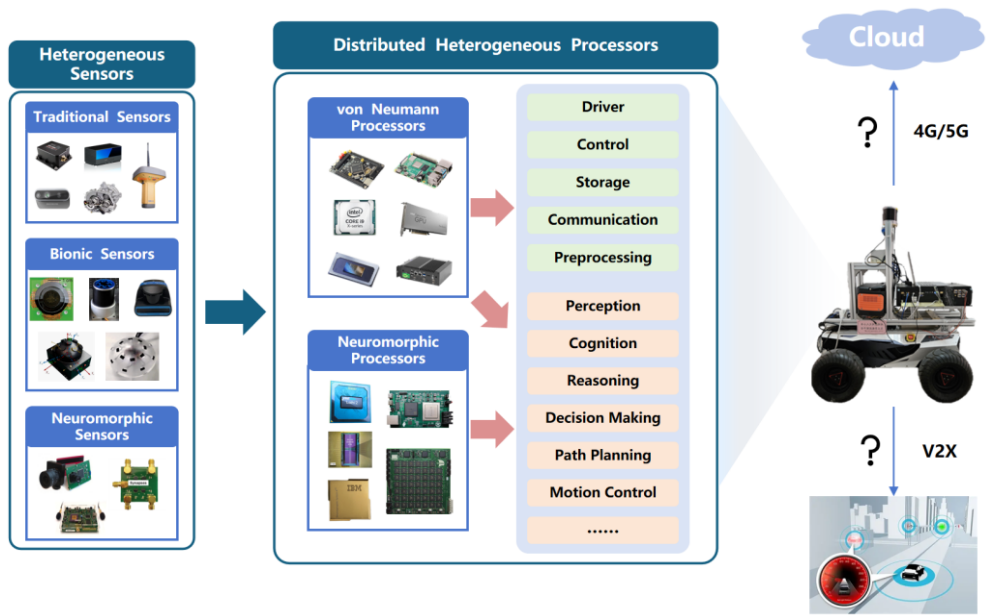

**Fig. 6** The architecture diagram of the proposed fusion pipeline.

The distributed processor framework depicted in Figure 6 encompasses a variety of von Neumann and neuromorphic processors. Machine PNT systems typically depend on MCUs, edge devices, and CPUs within the von Neumann architecture for functions such as communication, driver, control, storage, and signal preprocessing. Complex algorithm operations generally rely on CPUs/GPUs or AI-dedicated processors, while some simple algorithms can be implemented using TinyML. Neuromorphic processors, particularly heterogeneous chips like Tianjic and SpiNNaker2 that support the co-deployment of AI and brain-inspired paradigms, are well-suited for deploying certain algorithmic implementations at the functional level of PNT systems. This approach helps reduce power consumption and enhance computational efficiency. Additionally, when faced with limited computational and storage resources, neuromorphic processors can be paired with von Neumann processors.

In the context of heterogeneous PNT system deployment, the integration of heterogeneous sensors, edge AI chips, brain-inspired sensing-storage-computing chips/processors, and von Neumann processors should take into account a range of factors, including application objectives, system cost, power consumption, performance, and latency. Furthermore, as illustrated in Figure 6, this paper explores the potential cloud-edge collaboration and Vehicle-to-Everything (V2X) collaboration working modes for machine PNT systems with heterogeneous hardware integration.

## 4. Discussion and Perspectives

## 4.1 Global Optimum Outweighs Local Optimum

From a navigation standpoint, both machines and living organisms must acquire spatial cues from the physical world through their respective sensors. Machine systems rely on a diverse array of sensors including multimodal, biomimetic, brain-inspired, and even quantum sensors to perceive and harvest these cues. The data formats of these sensors include analog signals, digital signals, pulse signals, and hybrid signals, among others. To date, the long-standing industrial revolution of humanity has endowed manufacturing processes and material technologies with powerful capabilities, enabling a wide range of sensors to possess extremely high sensitivity and resolution. At present, even a basic, low-cost IMU can provide users with an angular measurement resolution at the level of $0.1°$ error, a level of precision that the spatial orientation perception systems of biological brains can hardly achieve.

In addition, machine systems can be equipped with different sensing devices in a targeted manner according to practical needs, and their sensing means are configurable with a wide range of choices. In contrast, living organisms rely solely on biological organs to perceive the environment, and the signals processed by different senses are all uniformly translated by the brain into expressions of physiological electrical signals. From the perspectives of pure sensitivity, resolution, dynamic range, and diversity, the sensory organs of most living beings in nature are almost completely outperformed by the high-precision sensors manufactured by humans. Moreover, for a healthy individual organism, once its growth and development are complete, it lacks the plasticity and configurability of machine sensing systems. However, the advantage of its biological sensing system lies in its natural unity, which does not face the issues of time synchronization between different sensors and loose/tight coupling of data usage that machine systems encounter.

Beyond sensors, the navigation algorithms employed by machine systems depend on a diverse array of processors, including CPUs, GPUs, AI accelerators, brain-inspired chips, quantum processors, and etc. Although neuromorphic computing has drawn extensively from neuroscience to develop advanced chips such as Tianjic and Loihi, these chips still fall significantly short of fully replicating the brain's general-purpose spatiotemporal generalization, in-memory computation, and self-organizing/adaptive characteristics, particularly in terms of energy efficiency and power consumption [103]. In contrast, the biological brain has evolved naturally, with its neural circuits handling spatial navigation through an architectural substrate that is entirely distinct from silicon. Machine systems fabricate chips from silicon using CMOS processes, while the brain utilizes neurons and synapses as its fundamental building blocks to form complex neural circuits. The latter achieves remarkable computational efficiency and robustness, with ultra-long-endurance and ultra-low-power operation that far surpasses any current artificial processor.

Of course, when comparing machines and living organisms as navigation platforms, one must also weigh their mechanical robustness against their agility. Machine systems, shaped by human industrial capability, far surpass flesh and bone in strength, yet they still lag behind the natural world in flexibility, environmental adaptability, autonomy, and energy efficiency. Even the most advanced robots, for example, remain unable to match the agility of a house cat.

In summary, apart from the brain's versatile cognitive intelligence and its ability to operate for long durations on minuscule power, human-engineered machines outperform biological systems on nearly every individual metric. Nevertheless, their overall navigational competence still falls short of that found in nature, especially in adaptability, robustness, and autonomy. Therefore, whether traditional or brain-inspired, PNT systems should strive for global optimum rather than local optimum: extracting the greatest navigational benefit at the lowest possible cost and energy expenditure. This, ultimately, is the answer that brain-inspired PNT must seek from the brain's spatial cognition intelligence.

## 4.2 GNSS Should Not Be Left Out

At present, many researchers label brain-inspired navigation as a GNSS-free autonomous technology. It is certainly true that endowing unmanned systems with highly autonomous and robust spatial cognition under GNSS-denied conditions is one of the goals brain-inspired navigation should pursue. Yet we do not endorse the deliberate exclusion of GNSS from brain-inspired PNT research, as such an approach would represent a strategic misstep.

Today, GNSS and other modern PNT technologies have already delivered mature, cost-effective solutions across civilian and industrial domains. When reliable and economical options like GNSS are available, bypassing them in favor of more convoluted alternatives is unwise. This does not imply that brain-inspired spatial cognition is unimportant. After all, humans, natural carriers of brain intelligence, routinely rely on smartphones and other smart devices to obtain location services for path planning and navigation, demonstrating that these approaches can coexist harmoniously. GNSS-based services have undeniably augmented human navigational capability. Therefore, mathematically grounded PNT and spatial cognition-driven brain-inspired PNT should be viewed as complementary partners, not competing adversaries.

Moreover, many migratory animals rely on globally invariant external cues, such as the stars, the geomagnetic field, gravity, and polarized light, to guide long-distance navigation [4]. These signals provide a unified spatiotemporal reference frame that enables the animals to determine their position and orientation throughout their journey. Insects also deposit pheromones in the environment for navigation or other life activities [104]. GNSS can be viewed as the "pheromone" humanity leaves during geodetic surveying: a public broadcast of space-time datum, spatial metrics, and geographic semantics for all human users.

When the spatial scale is vast, almost all animals, including humans, find it difficult to make optimal path planning in large-scale navigation tasks. The root cause is the absence of a "God's-eye view". Animal brains must rely on continuous integration of purely local sensory information. Consequently, long-range navigation in nature is almost never optimal [105]. We therefore argue that feeding GNSS's unified spatiotemporal datum ("God's-eye view") into the perception stage of a brain-inspired PNT system will markedly strengthen its spatial description model and path planning capabilities, especially when the spatial span is large.

Not only that, notably, our understanding of the brain's temporal coding, corresponding to the timing dimension, is limited.

This is particularly evident in phenomena such as time cells' neural functions and theta-gamma coupling. Additionally, engineered systems have an intrinsic requirement for precise time synchronization. These factors together prevent brain-inspired PNT from fully aligning its temporal representation with that of the brain. Therefore, brain-inspired PNT currently has significant deficiencies in the timing dimension that urgently need to be addressed. Since the timing dimension that is currently missing in brain-inspired PNT systems is exactly where GNSS PNT excels. Therefore, we advocate utilizing GNSS's timing capabilities to complement the missing timing dimension in brain-inspired PNT.

### 4.3 Building an Ecosystem and Evaluation Framework for Brain-Inspired PNT

At present, brain-inspired PNT remains in a stage of theoretical exploration and partial mechanism validation. Navigation platforms, datasets, and reference benchmarks are all incomplete, and a systematic evaluation framework is lacking. Current assessments rely on discrete metrics such as positioning accuracy and trajectory precision, which cannot capture the overall performance of brain-inspired PNT systems. Therefore, establishing a comprehensive evaluation framework, along with relevant technical standards, is essential to support a healthy research ecosystem and practical applications in this field.

Brain-inspired PNT seeks to replicate the core strengths of animal navigation, which include acquiring environmental knowledge, storing spatiotemporal memories, and rapidly retrieving these experiences to navigate complex terrains and achieve a goal with low power consumption and high robustness. For robots and other unmanned systems, the overriding concern is the integrated PNT capability. The key evaluation criteria should therefore be: the ability to learn autonomously and reason, to deliver efficient, intelligent navigation that meets a prescribed positioning accuracy while remaining low-power and highly robust, and to use these criteria to iteratively enrich and refine the evaluation framework itself.

Given that brain-inspired PNT intersects with traditional PNT across numerous technical dimensions, it is feasible to draw insights from the well-established, GNSS-centric PNT ecosystem, even in the absence of extensive brain-inspired case studies. We should utilize evaluation metrics that are common to both domains and subsequently enhance them with indicators that are unique to brain-inspired PNT. We believe that the brain-inspired PNT-specific increments compared to the traditional PNT evaluation system should focus on the simulation of the brain's spatial cognition intelligence, which is the most significant distinction between brain-inspired PNT and traditional PNT. Spatial cognition, however, is a vast domain. A critical question is how to define both the research scope and the capability boundaries of brain-inspired PNT at each developmental stage. Equally pressing is how to quantitatively assess a brain-inspired PNT system's spatial cognition competence. This entails designing quantitative rules and public benchmarks that can objectively measure spatial cognition ability within such systems.

Furthermore, brain-inspired PNT is also a cutting-edge intersection of AI, brain-inspired intelligence, and PNT technologies. As such, it can draw on the experiences of ecosystem and evaluation framework building from the AI and brain-inspired intelligence communities. Yet the roadmap of brain-inspired PNT has not yet solidified, and we cannot rule out the possibility that iterative development may give rise to entirely new directions. As a thought-provoking example, consider the potential introduction of brain-computer interfaces into brain-inspired PNT: a key theme in brain-inspired intelligence that could enable complementary and coordinated human-machine intelligence within PNT systems for harsh, deep space, deep ocean, polar, or other scenarios where conventional autonomy is unachievable. Thus, constructing the ecosystem and evaluation framework for brain-inspired PNT is a long-term, continuously evolving endeavor.

For the present stage, we recommend focusing on technical feasibility rather than absolute accuracy, because brain-inspired PNT is still in its infancy. Notably, its neuromorphic implementations already demonstrate non-negligible power and energy-efficiency advantages. Taking these factors together, we propose the following evaluation metrics for current-phase brain-inspired PNT.

① **Availability**: Design navigation scenarios and tasks with varying levels of complexity. Quantify availability as the success rate (expressed as a percentage) of task completion by the brain-inspired PNT system across diverse dynamic environments.

② **Accuracy**: From an end-to-end navigation perspective, measure accuracy by the final positional error, defined as the distance between the platform's actual arrival point and the intended goal (in meters), rather than relying on intermediate localization statistics.

③ **Energy Efficiency**: Power consumed per unit time while the brain-inspired PNT system performs the navigation task (watts).

④ **Efficiency**: Ratio of the path length actually travelled under brain-inspired PNT to the shortest feasible path pre-computed for the given scenario.

⑤ **Responsiveness**: Compute latency of the brain-inspired PNT algorithm under identical target scenarios, expressed as both average dynamic latency and maximum response delay.

Drawing on ecosystem-building practices in AI and brain-inspired intelligence communities, we propose the following layers for the brain-inspired PNT ecosystem:

① **Open-source frameworks and platforms**: brain-inspired PNT computation stacks, algorithms/models, tutorials, and application examples.

② **Datasets**: large-scale, multi-modal datasets collected in complex environments (event cameras, GNSS, RGB cameras, etc.) for training and evaluation.

③ **Hardware**: bionic robots, mobile robots, autonomous vehicles, bionic sensors, neuromorphic sensors/chips/modules.

④ **Standards, specifications and protocols**: ensure interoperability and compatibility among diverse PNT systems and models.

⑤ **Academic ecosystem and talent cultivation**: International conferences, forums, special issues, lectures, and programs aimed at cultivating strategic scientists and addressing the urgent need for specialists.

⑥ **Privacy and ethics**: formulate ethical guidelines and legal regulations to guarantee fairness, transparency, and security; reinforce data privacy and security when leveraging brain-inspired PNT big-data infrastructures.

⑦ **R&D community**: an active network of researchers, developers, and data scientists driving continuous innovation and improvement of brain-inspired PNT models and algorithms.

## 5. Concluding Remarks

To explore a fusion path between traditional high-precision PNT and brain-inspired PNT, this paper first systematically dissects the differences among traditional, biological, and brain-inspired PNT systems across the dimensions of principle, paradigm, and hardware. It points out that these pipelines are not in a zero-sum opposition, but complementary ecological niches. Building on this insight, we provide a feasible technical roadmap and an open-ecology blueprint for today's PNT technologies to transition from "tool-oriented" to "cognition-driven" paradigms. A preliminary fusion architecture, spanning observation, computation, decision, and hardware layers, is proposed and illustrated for reference.

Moreover, from a forward-looking perspective and grounded in our comparative analysis, we offer recommendations for the future development of brain-inspired PNT. We believe that only by deeply integrating the precision of traditional mathematical PNT with brain-inspired spatial cognition intelligence can we ultimately create a universal PNT system that is "brain-like" yet "brain-surpassing," meeting urgent demands for resilience, energy efficiency, and reliability in ubiquitous, complex, and even extreme scenarios.

## Reference

[1] Yang, G. Z., Bellingham, J., Dupont, P. E., Fischer, P., Floridi, L., Full, R., ... & Wood, R. (2018). The grand challenges of science robotics. *Science Robotics*, 3(14), eaar7650.

[2] Yin, B., Wang, Q., Zhang, P., Zhang, J., Wang, K., Wang, Z., ... & Fei-Fei, L. (2025). Spatial mental modeling from limited views. *arXiv preprint*, arXiv:2506.21458.

[3] Horton, T.W., Holdaway, R. N., Zerbini, A. N., Hauser, N., Garrigue, C., Andriolo, A., & Clapham, P. J. (2011). Straight as an arrow: humpback whales swim constant course tracks during long-distance migration. *Biology Letters*, 7(5), 674-679.

[4] Mouritsen. H. (2018). Long-distance navigation and magnetoreception in migratory animals. *Nature*, 558(7708), 50-59.

[5] Omer, D. B., Maimon, S. R., Las, L., & Ulanovsky, N. (2018). Social place-cells in the bat hippocampus. *Science*, 359(6372), 218-224.

[6] Finkelstein, A., Las, L., & Ulanovsky, N. (2016). 3-D maps and compasses in the brain. *Annual Review of Neuroscience*, 39(1), 171-196.

[7] Epstein, R. A., Patai, E. Z., Julian, J. B., & Spiers, H. J. (2017). The cognitive map in humans: spatial navigation and beyond. *Nature neuroscience*, 20(11), 1504-1513.

[8] Plag, H. P., Rothacher, M., Pearlman, M., Neilan, R., & Ma, C. (2009). The global geodetic observing system. *In Advances in Geosciences: Volume 13: Solid Earth (SE)* (pp. 105-127).

[9] Craig, J. J. (1989). Introduction to robotics: mechanics and control, 2nd Edition, Addison.

[10] Yao, Z., Qi, Y. & Lu, M. (2024). Subcarrier modulated navigation signal processing in GNSS: a review. *Satellite Navigation*, 5(1), 24.

[11] Maciejewska, A., & Maciuk, K. (2025). Research using GNSS (Global Navigation Satellite System) products – a comprehensive literature review. *Journal of Applied Geodesy*, 19(4), 555-574.

[12] Thrun, S., Burgard, W., & Fox, D. (2005). Probabilistic robotics (intelligent robotics and autonomous agents), MIT Press.

[13] Davison, A. C., & Huser, R. (2015). Statistics of extremes. *Annual Review of Statistics and its Application*, 2(1), 203-235.

[14] Schölkopf, B., Locatello, F., Bauer, S., Ke, N.R., Kalchbrenner, N., Goyal, A., & Bengio, Y. (2021). Toward causal representation learning. *Proceedings of the IEEE*, 109(5), 612-634.

[15] Ruderman, M.S., Iwasaki, M., & Chen, W. (2020). Motion-control techniques of today and tomorrow: a review and discussion of the challenges of controlled motion. *IEEE Industrial Electronics Magazine*, 14(1), 41-55.

[16] Low, I.I., & Giocomo, L.M. (2021). Fifty years of the brain's sense of space. *Nature*, 599(7885), 376-377.

[17] Kim, M., & Maguire, E.A. (2018). Encoding of 3D head direction information in the human brain. *Hippocampus*, 29(7), 619-629.

[18] McNaughton, B. L., Battaglia, F. P., Jensen, O., Moser, E. I., & Moser, M. B. (2006). Path integration and the neural basis of the 'cognitive map'. *Nature Reviews Neuroscience*, 7(8), 663-678.

[19] Giocomo, L. M. (2016). Environmental boundaries as a mechanism for correcting and anchoring spatial maps. *The Journal of physiology*, 594(22), 6501-6511.

[20] Rolls, E. T. (2020). Spatial coordinate transforms linking the allocentric hippocampal and egocentric parietal primate brain systems for memory, action in space, and navigation. *Hippocampus*, 30(4), 332-353.

[21] Alexander, A. S., Robinson, J. C., Stern, C. E., & Hasselmo, M. E. (2023). Gated transformations from egocentric to allocentric reference frames involving retrosplenial cortex, entorhinal cortex, and hippocampus. *Hippocampus*, 33(5), 465-487.

[22] Hinman, J. R., Chapman, G. W., & Hasselmo, M. E. (2019). Neuronal representation of environmental boundaries in egocentric coordinates. *Nature communications*, 10(1), 2772.

[23] Alexander, A. S., Carstensen, L. C., Hinman, J. R., Raudies, F., Chapman, G. W., & Hasselmo, M. E. (2020). Egocentric boundary vector tuning of the retrosplenial cortex. *Science advances*, 6(8), eaaz2322.

[24] Cheng, N., Dong, Q., Zhang, Z., Wang, L., Chen, X., & Wang, C. (2024). Egocentric processing of items in spines, dendrites, and somas in the retrosplenial cortex. *Neuron*, 112(4), 646-660.

[25] Behrens, T.E., Muller, T.H., Whittington, J.C., Mark, S., Baram, A.B., Stachenfeld, K.L., & Kurth-Nelson, Z. (2018). What is a cognitive map? Organizing knowledge for flexible behavior. *Neuron*, 100(2), 490-509.

[26] Ayala, A.P., Azuela, J.H., & Gutiérrez, A. (2008). Causal knowledge and reasoning by cognitive maps: pursuing a holistic approach. *Expert Systems with Applications*, 35(1-2), 2-18.

[27] Jiao, L., Ma, M., He, P., Geng, X., Liu, X., Liu, F., Ma, W., Yang, S., Hou, B., & Tang, X. (2024). Brain-inspired learning, perception, and cognition: a comprehensive review. *IEEE Transactions on Neural Networks and Learning Systems*, 36(4), 5921-5941.

[28] Lahr, M., & Donato, F. (2020). Navigation: how spatial cognition is transformed into action. *Current Biology*, 30(10), R430-R432.

[29] Olson, J.M., Li, J.K., Montgomery, S.E., & Nitz, D.A. (2019). Secondary motor cortex transforms spatial information into planned action during navigation. *Current Biology*, 30(10), 1845-1854.e4.

[30] Milford, M. J., Wyeth, G. F., & Prasser, D. (2004). RatSLAM: a hippocampal model for simultaneous localization and mapping. In *IEEE International Conference on Robotics and Automation, 2004. Proceedings. ICRA'04. 2004* (vol. 1, pp. 403-408). IEEE.

[31] Fleischer, J. G., Gally, J. A., Edelman, G. M., & Krichmar, J. L. (2007). Retrospective and prospective responses arising in a modeled hippocampus during maze navigation by a brain-based device. *Proceedings of the National Academy of Sciences*, 104(9), 3556-3561.

[32] Yang, Y., Bartolozzi, C., Zhang, H.H., & Nawrocki, R.A. (2023). Neuromorphic electronics for robotic perception, navigation and control: a survey.

Engineering Applications of Artificial Intelligence, 126, 106838.

[33] Milford, M., & Wyeth, G.F. (2008). Mapping a suburb with a single camera using a biologically inspired SLAM system. *IEEE Transactions on Robotics*, 24(5), 1038-1053.

[34] Milford, M., Wiles, J., & Wyeth, G.F. (2010). Solving navigational uncertainty using grid cells on robots. *PLoS Computational Biology*, 6(11), e1000995.

[35] Bai, Y., Shao, S., Zhang, J., Zhao, X., Fang, C., Wang, T., Wang, Y., & Zhao, H. (2024). A review of brain-inspired cognition and navigation technology for mobile robots. *Cyborg and Bionic Systems*, 5, 0128.

[36] Kanitscheider, I., & Fiete, I. (2017). Training recurrent networks to generate hypotheses about how the brain solves hard navigation problems. In *Advances in Neural Information Processing Systems* (pp. 4530-4539).

[37] Cueva, C. J., & Wei, X. X. (2018). Emergence of grid-like representations by training recurrent neural networks to perform spatial localization. In *the 6th International Conference on Learning Representations (ICLR)*. Vancouver, BC, Canada.

[38] Banino, A., Barry, C., Uria, B., Blundell, C., Lillicrap, T., Mirowski, P., ... & Kumaran, D. (2018). Vector-based navigation using grid-like representations in artificial agents. *Nature*, 557(7705), 429-433.

[39] Zhang, Y., Chen, Y., Zhang, J., Luo, X., Zhang, M., Qu, H., & Yi, Z. (2022). Minicolumn-based episodic memory model with spiking neurons, dendrites and delays. *IEEE transactions on neural networks and learning systems*, 35(5), 7072-7086.

[40] Tang, H., Yan, R., & Tan, K. C. (2017). Cognitive navigation by neuro-inspired localization, mapping, and episodic memory. *IEEE Transactions on Cognitive and Developmental Systems*, 10(3), 751-761.

[41] Wang, C., Wang, L., Lu, Z., Wu, C., Shou, G., & Wen, X. (2025). Autonomous driving via brain-inspired causality-aware contrastive learning with time-frequency prediction. *IEEE Internet of Things Journal*, 12(14), 26371-26386.

[42] Ball, D., Heath, S., Wiles, J., Wyeth, G., Corke, P., & Milford, M. (2013). OpenRatSLAM: an open source brain-based SLAM system. *Autonomous Robots*, 34(3), 149-176.

[43] Kaess, M., Ranganathan, A., & Dellaert, F. (2008). iSAM: incremental smoothing and mapping. *IEEE Transactions on Robotics*, 24(6), 1365-1378.

[44] Mur-Artal, R., & Tardós, J. D. (2017). ORB-SLAM2: an open-source slam system for monocular, stereo, and RGB-D cameras. *IEEE transactions on robotics*, 33(5), 1255-1262.

[45] Yu, F., Shang, J., Hu, Y., & Milford, M. (2019). NeuroSLAM: A brain-inspired SLAM system for 3D environments. *Biological cybernetics*, 113(5), 515-545.

[46] Steffen, L., da Silva, R. K., Ulbrich, S., Tieck, J. C. V., Roennau, A., & Dillmann, R. (2020, November). Networks of place cells for representing 3D environments and path planning. In *2020 8th IEEE RAS/EMBS International Conference for Biomedical Robotics and Biomechatronics (BioRob)* (pp. 1158-1165). IEEE.

[47] Liu, D., Lyu, Z., Zou, Q., Bian, X., Cong, M., & Du, Y. (2022). Robotic navigation based on experiences and predictive map inspired by spatial cognition. *IEEE/ASME Transactions on Mechatronics*, 27(6), 4316-4326.

[48] Paredes-Vallés, F., Hagenaars, J. J., Dupeyroux, J., Stroobants, S., Xu, Y., & de Croon, G. C. (2024). Fully neuromorphic vision and control for autonomous drone flight. *Science Robotics*, 9(90), eadi0591.

[49] Novo, A., Lobon, F., Garcia de Marina, H., Romero, S., & Barranco, F. (2024). Neuromorphic perception and navigation for mobile robots: a review. *ACM Computing Surveys*, 56(10), 246.

[50] Wu, S., Jiang, T., Zhang, G., Schoenemann, B., Neri, F., Zhu, M., ... & Kuhnert, K. D. (2017). Artificial compound eye: a survey of the state-of-the-art. *Artificial Intelligence Review*, 48(4), 573-603.

[51] Vanarse, A., Osseiran, A., & Rassau, A. (2016). A review of current neuromorphic approaches for vision, auditory, and olfactory sensors. *Frontiers in Neuroscience*. 10, 115.

[52] Li, S., Kong, F., Xu, H., Guo, X., Li, H., Ruan, Y., ... & Guo, Y. (2023). Biomimetic polarized light navigation sensor: a review. *Sensors*, 23(13), 5848.

[53] Jung, Y. H., An, J., Hyeon, D. Y., Wang, H. S., Kim, I., Jeong, C. K., ... & Lee, K. J. (2024). Theoretical basis of biomimetic flexible piezoelectric acoustic sensors for future customized auditory systems. *Advanced Functional Materials*, 34(10), 2309316.

[54] Liu, S., Schaik, A.V., Minch, B.A., & Delbrück, T. (2014). Asynchronous binaural spatial audition sensor with 2 × 64 × 4 channel output. *IEEE Transactions on Biomedical Circuits and Systems*, 8(4), 453-464.

[55] Xie, S., Lu, Y., Zhang, S., Wang, L., & Zhang, X. (2010). Electro-optical gas sensor based on a planar light-emitting electrochemical cell microarray. *Small*, 6(17), 1897-1899.

[56] Loutfi, A., Coradeschi, S., Karlsson, L., & Broxvall, M. (2004, September). Putting olfaction into action: using an electronic nose on a multi-sensing mobile robot. *In 2004 IEEE/RSJ International Conference on Intelligent Robots and Systems (IROS)* (vol. 1, pp. 337-342). IEEE.

[57] Kim, C., Lee, K. K., Kang, M. S., Shin, D. M., Oh, J. W., Lee, C. S., & Han, D. W. (2022). Artificial olfactory sensor technology that mimics the olfactory mechanism: a comprehensive review. *Biomaterials Research*, 26(1), 40.

[58] Imam, N., & Cleland, T.A. (2019). Rapid online learning and robust recall in a neuromorphic olfactory circuit. *Nature Machine Intelligence*, 2(3), 181-191.

[59] Han, J. K., Kang, M., Jeong, J., Cho, I., Yu, J. M., Yoon, K. J., ... & Choi, Y. K. (2022). Artificial olfactory neuron for an in-sensor neuromorphic nose. *Advanced Science*, 9(18), 2106017.

[60] Navaraj, W., & Dahiya, R. (2019). Fingerprint-enhanced capacitive-piezoelectric flexible sensing skin to discriminate static and dynamic tactile stimuli. *Advanced Intelligent Systems*, 1(7), 1900051.

[61] Wang, X., Dong, L., Zhang, H., Yu, R., Pan, C., & Wang, Z. (2015). Recent progress in electronic skin. *Advanced Science*, 2(10), 1500169.

[62] Ward-Cherrier, B., Pestell, N., & Lepora, N.F. (2020). NeuroTac: a neuromorphic optical tactile densor applied to texture recognition. *In 2020 IEEE International Conference on Robotics and Automation (ICRA)* (pp. 2654-2660). IEEE.

[63] Zheng, K., Qian, K., Woodford, T., & Zhang, X. (2023). Neuroradar: a neuromorphic radar sensor for low-power IoT systems. *In Proceedings of the 21st ACM Conference on Embedded Networked Sensor Systems* (pp. 223-236).

[64] Jouni, Z., Soupizet, T., Wang, S., Benlarbi-Delai, A., & Ferreira, P.M. (2023). RF neuromorphic spiking sensor for smart IoT devices. Analog Integrated Circuits and Signal Processing, 117, 3-20.

[65] Wang, F., & Chen, M. (2024). Bionic study of distance-azimuth discrimination of multi-scattered point objects in bat bio-sonar. *Bioinspiration & Biomimetics*, 19(2), 026011.

[66] Cheng, Q., Ge, Y., Lin, B., Zhou, L., Mao, H., & Zhao, J. (2024). Capacitive bionic magnetic sensors based on one-step biointerface preparation. *ACS Applied Materials & Interfaces*, 16(6), 6789-6798.

[67] Introducing the Colossus? MK2 GC200 IPU. graphcore.ai. https://www.graphcore.ai/products/ipu.

[68] Tang, G., Shah, A., & Michmizos, K. P. (2019, November). Spiking neural network on neuromorphic hardware for energy-efficient unidimensional SLAM. In *2019 IEEE/RSJ International Conference on Intelligent Robots and Systems (IROS)* (pp. 4176-4181). IEEE.

[69] Yoon, J. H., & Raychowdhury, A. (2020). NeuroSLAM: a 65-nm 7.25-to-8.79-TOPS/W mixed-signal oscillator-based SLAM accelerator for edge robotics. *IEEE Journal of Solid-State Circuits*, 56(1), 66-78.

[70] Pei, J., Deng, L., Song, S., Zhao, M., Zhang, Y., … & Shi, L. (2019). Towards artificial general intelligence with hybrid Tianjic chip architecture. *Nature*, 572(7767), 106-111.

[71] Gonzalez, H., Huang, J., Kelber, F., Nazeer, K., … & Mayr, C. (2024). SpiNNaker2: a large-scale neuromorphic system for event-based and asynchronous machine learning. *ArXiv*, abs/2401.04491.

[72] Rathi, N., Chakraborty, I., Kosta, A., Sengupta, A., Ankit, A., Panda, P., & Roy, K. (2023). Exploring neuromorphic computing based on spiking neural networks: algorithms to hardware. *ACM Computing Surveys*, 55(12), 1-49.

[73] He, X., Mo, L., & Wang, Q. (2023). An attention-assisted UWB ranging error compensation algorithm. *IEEE Wireless Communications Letters*, 12(3), 421-425.

[74] Kannapiran, S., Bendapudi, N., Yu, M. Y., Parikh, D., Berman, S., Vora, A., & Pandey, G. (2023). Stereo visual odometry with deep learning-based point and line feature matching using an attention graph neural network. In *2023 IEEE/RSJ International Conference on Intelligent Robots and Systems (IROS)* (pp. 3491-3498). IEEE.

[75] Yang, N., Stumberg, L. V., Wang, R., & Cremers, D. (2020). D3VO: deep depth, deep pose and deep uncertainty for monocular visual odometry. In *Proceedings of the IEEE/CVF conference on computer vision and pattern recognition* (pp. 1281-1292).

[76] Brossard, M., Barrau, A., & Bonnabel, S. (2020). AI-IMU dead-reckoning. *IEEE Transactions on Intelligent Vehicles*, 5(4), 585-595.

[77] Du, N., Huang, Y., Dai, A. M., Tong, S., Lepikhin, D., Xu, Y., ... & Cui, C. (2022). Glam: efficient scaling of language models with mixture-of-experts. In *Proceedings of the 39th International Conference on Machine Learning (ICML)* (Vol. 162, pp. 5547-5569). PMLR.

[78] Yu, F., Wu, Y., Ma, S., Xu, M., Li, H., Qu, H., ... & Shi, L. (2023). Brain-inspired multimodal hybrid neural network for robot place recognition. *Science Robotics*, 8(78), eabm6996.

[79] Pellerito, R., Cannici, M., Gehrig, D., Belhadj, J., Dubois-Matra, O., Casasco, M., & Scaramuzza, D. (2024, October). Deep visual odometry with events and frames. In *2024 IEEE/RSJ International Conference on Intelligent Robots and Systems (IROS)* (pp. 8966-8973). IEEE.

[80] Gong, G., Hu, F., Wang, F., Muddassir, M., Zhou, P., Li, L., ... & Navarro-Alarcon, D. (2025). TEVIO: thermal-aided event-based visual inertial odometry for robust state estimation in challenging environments. *IEEE Transactions on Instrumentation and Measurement*, 74, 1-11.

[81] Jacobson, A., Chen, Z., & Milford, M. (2018). Leveraging variable sensor spatial acuity with a homogeneous, multi-scale place recognition framework. *Biological Cybernetics*, 112(3), 209-225.

[82] Liu, X., Chen, L., Jiao, Z., Yu, F., Lu, X., Liu, Z., & Ruan, Y. (2023). A neuro-inspired positioning system integrating MEMS sensors and DTMB signals. *IEEE Transactions on Broadcasting*, 69(3), 823-831.

[83] Yang, C., Xiong, Z., Liang, X., & Liu, J. (2024). Brain-inspired multimodal navigation with multiscale hippocampal-entorhinal neural network. *IEEE Transactions on Instrumentation and Measurement*, 73, 1-17.

[84] Liu, D., Lyu, Z., Li, B., Zou, Q., Du, Y., Bian, X., & Cong, M. (2021, July). Semi-Bionic SLAM Based on Visual Odometry and Deep Learning Network. In *2021 IEEE 11th Annual International Conference on CYBER Technology in Automation, Control, and Intelligent Systems (CYBER)* (pp. 293-299). IEEE.

[85] Li, J., Tang, H., & Yan, R. (2022). A hybrid loop closure detection method based on brain-inspired models. *IEEE Transactions on Cognitive and Developmental Systems*, 14(4), 1532-1543.

[86] Iodice, G. M. (2022). TinyML Cookbook: Combine artificial intelligence and ultra-low-power embedded devices to make the world smarter. Packet Publishing Ltd.

[87] Li, X., Chen, X., Guo, R., Wu, Y., Zhou, Z., Yu, F., & Lu, H. (2025). NeuroVE: brain-inspired linear-angular velocity estimation with spiking neural networks. *IEEE Robotics and Automation Letters*, 10(3), 2375-2382.

[88] MacDonald, C. J., Lepage, K. Q., Eden, U. T., & Eichenbaum, H. (2011). Hippocampal "time cells" bridge the gap in memory for discontiguous events. *Neuron*, 71(4), 737-749.

[89] Shimbo, A., Izawa, E. I., & Fujisawa, S. (2021). Scalable representation of time in the hippocampus. *Science Advances*, 7(6), eabd7013.

[90] Han, Y., Huang, G., Song, S., Yang, L., Wang, H., & Wang, Y. (2021). Dynamic neural networks: a survey. *IEEE Transactions on Pattern Analysis and Machine Intelligence*, 44(11), 7436-7456.

[91] Guo, J., Chen, C. P., Liu, Z., & Yang, X. (2024). Dynamic neural network structure: a review for its theories and applications. *IEEE Transactions on Neural Networks and Learning Systems*, 36(3), 4246-4266.

[92] Xu, T., & Barak, O. (2020). Implementing inductive bias for different navigation tasks through diverse RNN attractors. *arXiv preprint arXiv:2002.02496*.

[93] Li, H., Hu, G., Liu, S., Ma, M., Chen, Y., & Zhao, D. (2025). NeuronsGym: a hybrid framework and benchmark for robot navigation with sim2real policy learning. *IEEE Transactions on Emerging Topics in Computational Intelligence*, 9(3), 2491-2505.

[94] Liu, Q., Wang, G., Liu, Z., & Wang, H. (2025). Visuomotor navigation for embodied robots with spatial memory and semantic reasoning cognition. *IEEE Transactions on Neural Networks and Learning Systems*, 36(5), 9512-9523.

[95] Gao, F., Tang, J., Wang, J., Li, S., & Yu, J. (2024). Enhancing scene understanding for vision and language navigation by knowledges awareness. *IEEE Robotics and Automation Letters*, 9(12), 10874-10881.

[96] Langdon, A., Botvinick, M., Nakahara, H., Tanaka, K., Matsumoto, M., & Kanai, R. (2022). Meta-learning, social cognition and consciousness in brains and machines. *Neural Networks*, 145, 80-89.

[97] Zhuang, F., Qi, Z., Duan, K., Xi, D., Zhu, Y., Zhu, H., ... & He, Q. (2021). A comprehensive survey on transfer learning. *Proceedings of the IEEE*, 109(1), 43-76.

[98] Parisi, G. I., Kemker, R., Part, J. L., Kanan, C., & Wermter, S. (2019). Continual lifelong learning with neural networks: a review. *Neural networks*, 113, 54-71.

[99] Schölkopf, B., Locatello, F., Bauer, S., Ke, N. R., Kalchbrenner, N., Goyal, A., & Bengio, Y. (2021). Toward causal representation learning. *Proceedings of the IEEE*, 109(5), 612-634.

[100] Li, W., Qin, C., Zhang, T., Mao, X., ... & Jiao, L. (2024). Review: brain-inspired intelligent navigation modeling technology and its application. *Systems Engineering and Electronics*, 46(11), 3844-3861.

[101] X. Dong, Z. Yu, W. Cao, Y. Shi & Q. Ma. (2019). A survey on ensemble learning. *Frontiers of Computer Science*, 14(5), 241-258.

[102] Bartolozzi, C., Indiveri, G., & Donati, E. (2022). Embodied neuromorphic intelligence. *Nature communications*, 13(1), 1024.

[103] Zhang, W., Gao, B., Tang, J., Yao, P., Yu, S., Chang, M. F., ... & Wu, H. (2020). Neuro-inspired computing chips. *Nature electronics*, 3(7), 371-382.

[104] Heinze, S., Narendra, A., & Cheung, A. (2018). Principles of insect path integration. *Current Biology*, 28(17), R1043-R1058.

[105] Bongiorno, C., Zhou, Y., Kryven, M., Theurel, D., Rizzo, A., Santi, P., ... & Ratti, C. (2021). Vector-based pedestrian navigation in cities. *Nature computational science*, 1(10), 678-685.